%% file: kws_interspeech2021.tex
\documentclass[a4paper]{article}

\usepackage{INTERSPEECH2021}

\usepackage{url}
\usepackage{authblk}
\usepackage{hyperref}
\usepackage{tabularx}
\usepackage{booktabs}
\usepackage{graphicx}
\usepackage{amssymb,amsmath,bm}
\usepackage{algorithm}
\usepackage{algpseudocode}
\usepackage{textcomp}
\usepackage{comment}
\usepackage{multirow}
\usepackage{caption}
\usepackage{subfigure}

\def\vec#1{\ensuremath{\bm{{#1}}}}

\input{math_commands.tex}

\title{Exploring Machine Speech Chain for Domain Adaptation and Few-Shot Speaker Adaptation}

\usepackage{blindtext}
\makeatletter
\def\name#1{\gdef\@name{#1\\}}
\makeatother

\name{{\textit{Fengpeng Yue$^1$$^*$\thanks{* Work done during internship at Microsoft China.}, Yan Deng$^2$, Lei He$^2$,  Tom Ko$^1$$^{\dag}$\thanks{\dag corresponding author.}}}}

\address{$^{\1}$Department of Computer Science and Engineering,  Southern University of Science and Technology, Shenzhen, China
$^2$Microsoft China}
\email{11930381@mail.sustech.edu.cn, \{yaden,helei\}@microsoft.com, tomkocse@gmail.com}

\begin{document}

\maketitle
\begin{abstract}
Machine Speech Chain, which integrates both end-to-end (E2E) automatic speech recognition (ASR) and text-to-speech (TTS) into one circle for joint training, has been proven to be effective in data augmentation by leveraging large amounts of unpaired data. In this paper, we explore the TTS→ASR pipeline in speech chain to do domain adaptation for both neural TTS and E2E ASR models, with only text data from target domain. We conduct experiments by adapting from audiobook domain (LibriSpeech) to presentation domain (TED-LIUM), there is a relative word error rate (WER) reduction of 10$\%$ for the E2E ASR model on the TED-LIUM test set, and a relative WER reduction of 51.5$\%$ in synthetic speech generated by neural TTS in the presentation domain. Further, we apply few-shot speaker adaptation for the E2E ASR by using a few utterances from target speakers in an unsupervised way, results in additional gains.


\end{abstract}

\section{Introduction}
\label{sec:intro}
E2E ASR has shown great potential and achieved start-of-the-art performance in a lot of scenarios.
The three commonly used methods include connectionist temporal classification (CTC) \cite{graves2006connectionist}, recurrent neural network Transducer (RNN-T) \cite{graves2012sequence} and attention-based encoder-decoder (AED) \cite{chan2015listen}.
On the other hand, neural approaches have become very popular in the field of TTS
\cite{ren2019fastspeech,shen2018natural}. 
The speech generated by start-of-the-art models are very difficult to be distinguished from human speech. 
Although ASR and TTS have both achieved breakthrough in a lot of benchmarks, they still suffer from data mismatch in real life applications.
Therefore, adaptation techniques are still needed to tackle the data mismatch problem.

Text-domain mismatch
is a common cause of performance degradation in ASR and TTS. 
For hybrid ASR, the acoustic model and language model are trained separately, text-domain adaptation or machine translation models can be used on language model to improve the performance in target domain \cite{cucu2012asr,cucu:hal-00959159}.
For E2E ASR, text-domain adaptation can be achieved by fusing additional language models into language-independent sequence-to-sequence ASR via transfer learning \cite{inaguma2019transfer}.
In parametric TTS, an approach for generating optimized script for adaptation is proposed in \cite{chu2002domain}.
For attention-based neural TTS, it is still a challenge to get good attention for unseen text, which results in incorrect pronunciation,skipping or repetition issues.
\cite{he2019robust} proposed a stepwise monotonic attention method to improve the robustness of the out-of-domain text.

Speaker adaptation aims at reducing the mismatch caused by speaker variability.
In ASR, speaker adaptation techniques can be divided into three categories \cite{delcroix2018auxiliary}, namely model retraining \cite{neto1995speaker,swietojanski2014learning}, feature transformation \cite{gales1998maximum,seide2011feature} and auxiliary feature-based adaptation \cite{saon2013speaker,delcroix2018auxiliary}.
Model retraining methods focus on fine-tuning the model parameters and they usually require a certain amount of adaptation data.
FMLLR \cite{gales1998maximum} and VTLN \cite{cohen1995vocal} are two typical feature transformation methods.
Both of them exploit a separately trained Gaussian mixture model to compute the feature transformation.
Auxiliary feature-based adaptation usually uses speaker characteristic as part of the input of the acoustic model. 
It can be applied to both hybrid and E2E ASR.

Recently, machine speech chain \cite{tjandra2017listening} has drawn  a lot of research interests.
It provides a feasible way to train ASR and TTS models with unpaired data. 
Furthermore, it can improve the two models concurrently by allowing them to teach each other.
Most existing adaptation methods adapt either ASR or TTS model individually and do not consider the advantage claimed by machine speech chain.
This motivates us to investigate on adaptation with the machine speech chain framework.

In this paper, we first explore the machine speech chain model for text-domain adaptation.
We adopt the TTS→ASR pipeline to build the adaptation framework, and conduct experiments by adapting from audiobook domain (LibriSpeech \cite{panayotov2015librispeech}) to presentation domain (TED-LIUM \cite{rousseau2012ted}).
In contrast to existing domain adaptation methods, which only helps ASR learning target text-domain knowledge, our approach can also help ASR learning corresponding acoustic representation by generating speech features of the texts in target domain using TTS.
At the same time, the TTS model is also adapted to the target domain by using ASR to provide a feedback signal on the quality of synthetic features.
Experiments show that our approach can significantly improve the performance of TTS on the target text-domain. 
Much fewer attention errors are observed.
After joint tranining, there is a relative WER reduction of 10$\%$ for the E2E ASR model on the target domain test set, and a relative WER reduction of 51.5$\%$ in synthetic speech generated by neural TTS for input texts from target domain.

We further apply speaker adaptation to the ASR model on top of domain adaptation in the machine speech chain framework.
We use TTS to generate speech features of target speakers for texts from target domain, to adapt ASR model.
Results show that our way of speaker adaptation achieves a further 2.6$\%$ relative WER reduction for ASR.
Similar idea has been proposed in \cite{tjandra2018machine}, which uses one-shot speaker adaptation to leverage more unseen speakers, to further improve both TTS and ASR. Here, we try to extend one-shot to few-shot and show the impact on ASR, by adapting to a pool of target speakers in an unsupervised way.

The rest of the paper is organized as follows. In Section 2 we review the machine speech chain framework. In Section 3 we introduce our adaptation approach. In Section 4 we describe the details of our experiments. Section 5 is the conclusion and future work.

\section{Machine Speech Chain}
\label{sec:maml}

The machine speech chain is built based on the interactive learning process of human speech perception and generation. It consists of both TTS and ASR, which are trained jointly to improve each other.
The ASR task is to transcribe speech into text, similar to the human speech perception. TTS is a speech generation task, which generates speech for any given text.

In the machine speech chain, attention-based E2E ASR and  neural TTS are used. Given the input speech $X$, E2E ASR directly models the conditional probability $P(Y|X)$ between speech $X$ and the token sequence $Y=\left\{y_1,y_2,...y_T\right\}$ of length T:
\begin{equation}
p_{ASR}(Y|X)= \prod_{i}^{T}p(y_i|y_1,y_2,...y_{i-1},X)
\end{equation}
where token $y_i$ can be character, subword or word.
The cross entropy loss is used to optimize ASR model.The loss function is as follows:
\begin{equation}
\begin{aligned}
    L_{ASR}(Y^*,X)&=-logp_{ASR}(Y^{*}|X) \\
&=-\sum_{i}^{T}logp_{ASR}(y_i^{*}|y_1^{*},y_2^{*},...y_{i-1}^{*},X)
\end{aligned}
\end{equation}
where $y_i^{*}$ is the ground-truth of token.

TTS uses an attention-based E2E model similar to ASR. 
It predicts the spectrogram for input text and calculates three kinds of losses between prediction and groundtruth: mean square error(MSE), mean absolute error(MAE), and stop token binary cross-entropy(BCE).
\begin{equation}
\begin{aligned}
    L_{TTS}&=L_{MSE}+L_{MAE}+L_{BCE}
\end{aligned}
\end{equation}
MSE loss can be interpreted as the minus log-probability of the speech features for Gaussian and Laplace distributions $-logP_{TTS}(X|Y)$ for constant scale parameters \cite{baskar2019semi}:
\begin{equation}
\begin{aligned}
-log p_{TTS}(X^{*}|Y) \propto ||X^{*}- \hat X||^2
\end{aligned}
\end{equation}
Where $X^{*}$ and $Y$ are groundtruth of spectrogram and input text token, respectively. $\hat X$ is the predicted spectrogram of TTS.

There are two pipelines in speech chain: one is TTS→ASR to leverage unpaired text and the other is ASR→TTS to leverage unpaired speech.
Only TTS→ASR pipeline is needed for domain adaptation when only text data are available.

In the TTS→ASR pipeline, 
TTS generates speech spectrogram $\hat X$ for given unpaired text $Y$.
\begin{equation}
\hat X=arg\mathop{max}\limits_{X} \left\{p_{TTS}(X|Y) \right\} \\
\end{equation}
Then $Y$ and $\hat X$ are used to calculate $L_{TTS\rightarrow ASR}$  with the ASR model:

\begin{equation}
L_{TTS\rightarrow ASR}=L_{ASR}(\hat X,Y)
\end{equation}

There have been some studies on using TTS to do data augmentation  to improve the performance of ASR.
In \cite{rossenbach2020generating}, TTS and ASR are trained separately, which can avoid the effort of training new models using the same acoustic feature for joint training.
Its experiment verified that even for the ASR model that is trained on 960 hours of data with SpecAugment \cite{48482}, there is still slight improvement after adding synthetic data generated by TTS.
With the help of consistency loss between ASR prediction on real and synthetic speech, the paired training data for ASR can be reduced by half with little degradation in performance \cite{wang2020improving}.
In addition to using the TTS→ASR pipeline to improve the performance of ASR, \cite{liu2018improving} uses it to resolve the robustness issue for neural TTS, which is induced by using a reference with totally different content for style transfer.

\begin{figure}[th]
\centering
\includegraphics[height=4cm,width=8cm]{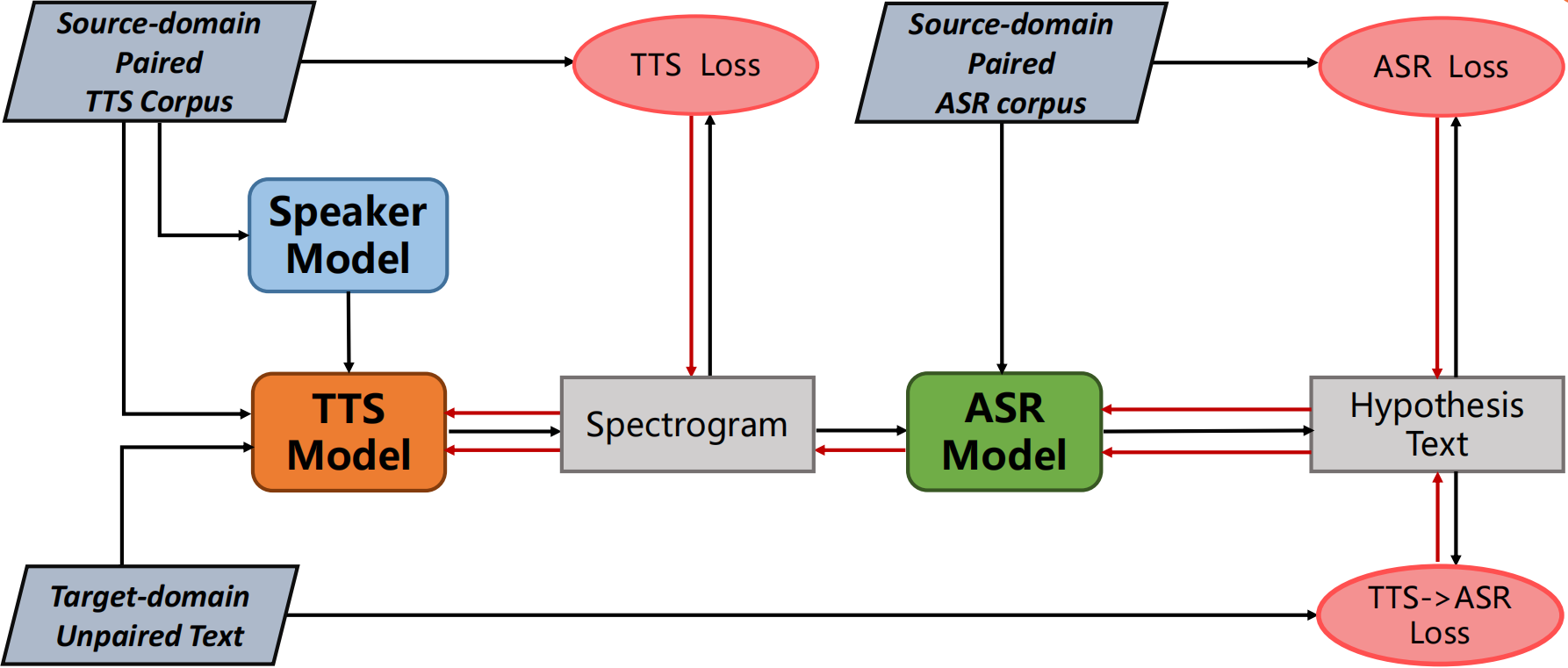}
\caption{Joint training framework. The black line represents forward calculation, and the red line represents back propagation. 
The gradients generated by TTS and ASR loss are only used to update their respective models, while TTS→ASR loss is used to update both models at the same time. 
}
\label{fig1}
\end{figure}
\section{Adaptation Framework}
For text-domain adaptation, only the unpaired text data is available.
In contrast to most prior works which usually train the ASR or TTS individually, we explore updating the ASR and TTS parameters simultaneously in the TTS→ASR pipeline\footnote{The codes are based on ESPnet \cite{watanabe2018espnet} and available at https://github.com/fengpeng-yue/ASRTTS}.
We first pretrain the ASR and TTS models with the source domain data, and then use TTS→ASR pipeline to do joint training for domain adaptation, as shown in Figure \ref{fig1}.
In joint training, we use source domain paired data to calculate ASR and TTS loss and use target domain unpaired text data to calculate TTS→ASR loss.
The unpaired text is randomly matched with all speakers in training corpus when TTS→ASR loss is calculated. 
The total loss can be expressed as follows:
\begin{equation}
L_{txt\_adapt} = L_{TTS} + L_{ASR} + L_{TTS \rightarrow{ASR}}
\end{equation}

We observe that $L_{TTS \rightarrow{ASR}}$ is having a larger dynamic range than $L_{TTS}$.
Thus, the gradient of $L_{TTS \rightarrow{ASR}}$ is much larger than those of $L_{TTS}$ when they arrive at the TTS model in the backpropagation.
To balance the influence of the two losses, we first perform gradient clipping on the gradients, and then accumulate gradient to update the TTS.
The $L_{ASR}$ and $L_{TTS}$ only update ASR and TTS, respectively.

We then further apply few-shot speaker adaptation on the ASR and keep the TTS model fixed.
As shown in Figure \ref{fig1}, we use the speaker model \cite{snyder2018x} pretrained on two large speaker verification datasets \cite{nagrani2017voxceleb,chung2018voxceleb2}.
The speaker embedding extracted by the speaker model is  concatenated to the output hidden state of the TTS encoder. 
To verify the effectiveness of few-shot speaker adaptation for ASR, we randomly select 1 or 5 speaker embeddings from each target speaker in test set.
The loss used to update ASR is as follow:
\begin{equation}
L_{spk\_adapt} = L_{ASR} + L_{TTS \rightarrow{ASR}}
\end{equation}


\begin{figure*}
\centering
\subfigure{
\begin{minipage}[t]{0.33\linewidth}
\centering
\includegraphics[width=2in]{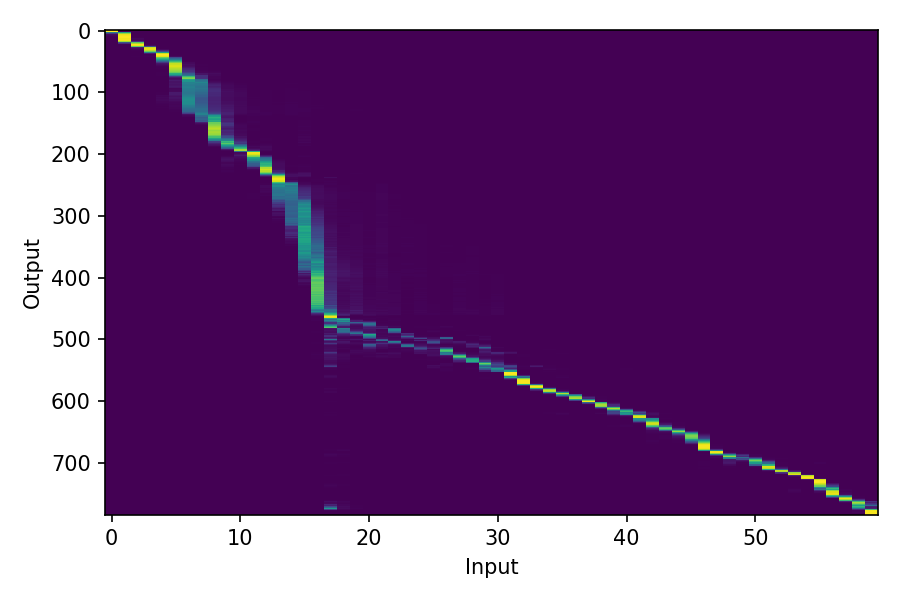}
\end{minipage}%
}%
\subfigure{
\begin{minipage}[t]{0.33\linewidth}
\centering
\includegraphics[width=2in]{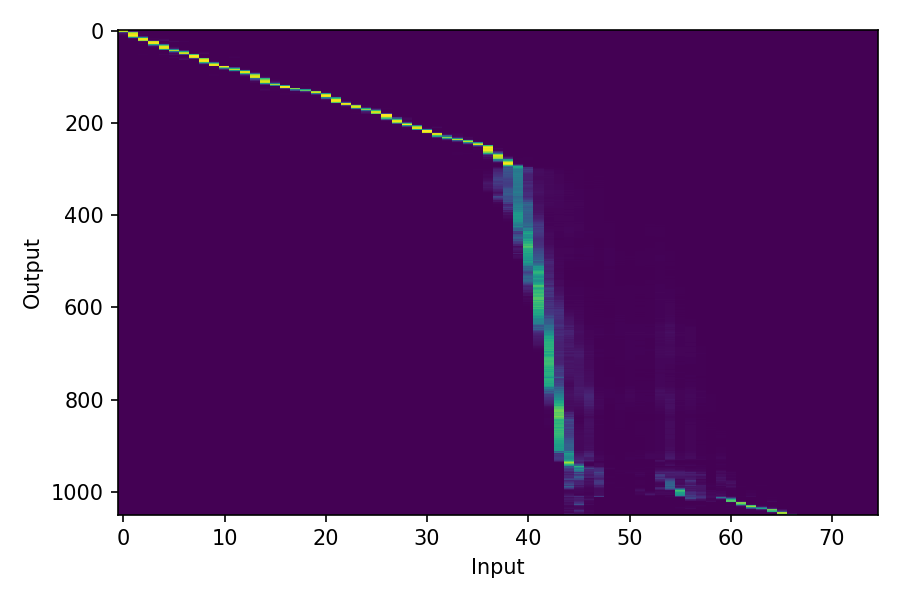}
\caption*{\normalfont (a) Baseline}
\end{minipage}%
}%
\subfigure{
\begin{minipage}[t]{0.33\linewidth}
\centering
\includegraphics[width=2in]{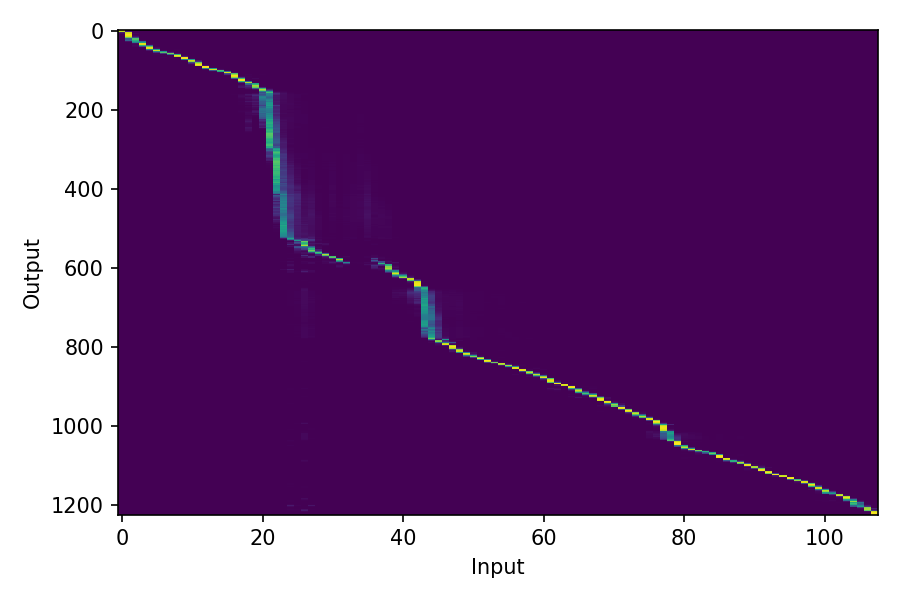}
\end{minipage}%
}%

\subfigure{
\begin{minipage}[t]{0.33\linewidth}
\centering
\includegraphics[width=2in]{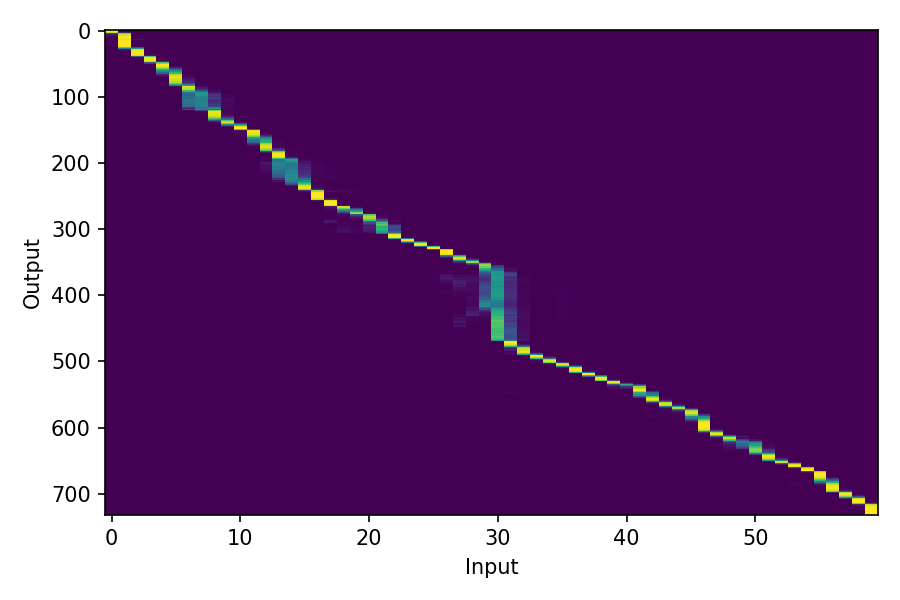}
\end{minipage}%
}%
\subfigure{
\begin{minipage}[t]{0.33\linewidth}
\centering
\includegraphics[width=2in]{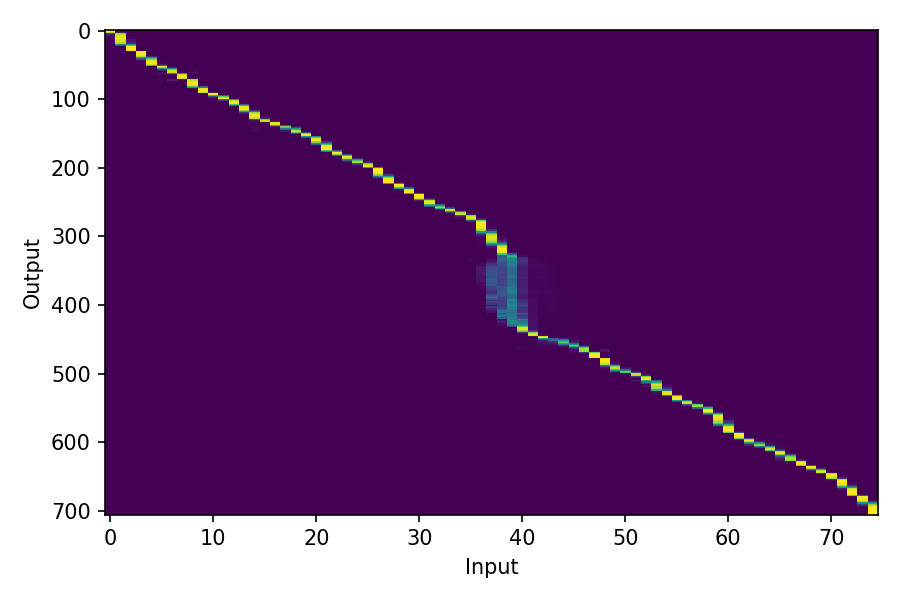}
\caption*{\normalfont (b) Joint training}
\end{minipage}%
}%
\subfigure{
\begin{minipage}[t]{0.33\linewidth}
\centering
\includegraphics[width=2in]{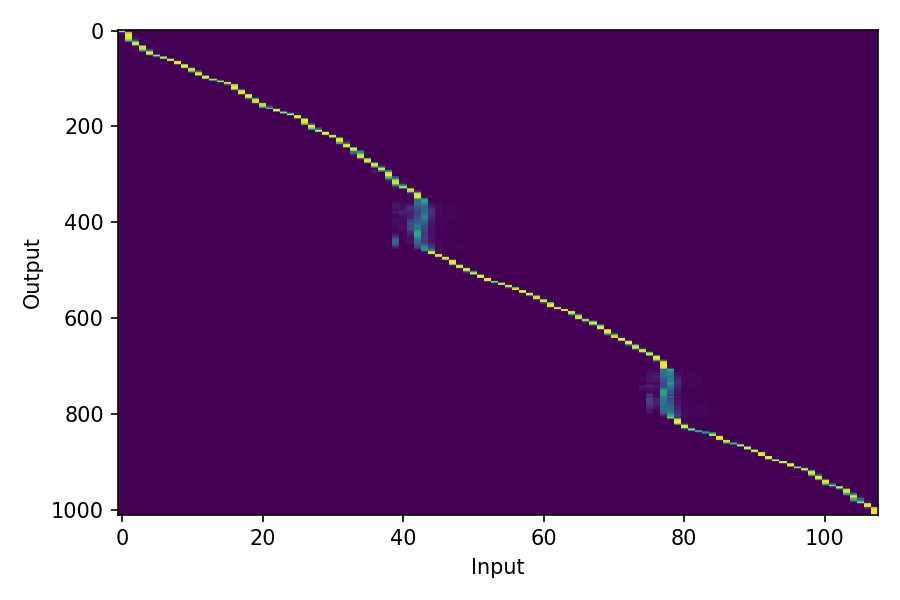}
\end{minipage}%
}%
\caption{Attention alignments on out-of-domain text. The first and the second rows represent the alignments of the same text synthesized by the baseline and the jointly trained system. The abscissa represents encoder timesteps, and the ordinate represents decoder timesteps.}
\label{fig2}
\end{figure*}
\section{Experiment}
\label{sec:exp}

\subsection{Experimental Setup}

\subsubsection{Dataset}
We choose LibriSpeech \cite{panayotov2015librispeech} as ASR source domain dataset and LribriTTS \cite{zen2019libritts} as TTS source domain dataset.
We pretrain our ASR baseline model on LibriSpeech train-clean-460 set and pretrain our TTS baseline model on LibriTTS train-clean-460 set.
To make the spectrogram synthesized by TTS in joint training consistent with the sampling frequency of the pretrained ASR model, we downsample LibriTTS to 16kHZ. 
We use 128-dimension Mel-filter bank coefficients (FBANK) features computed on 50ms window with 10ms shift.
We use character as ASR token units and phone as TTS token units. 

The text contents of LiriSpeech and LibriTTS come from audiobook, which are in written style with formal wording and strict grammar.
TED-LIUM text comes from different people’s presentation, which are in spoken style and more casual grammatically.
Therefore, we choose it as the target domain and use all texts from training set for joint training with TTS→ASR pipeline.
To match the pretrained TTS model, we add punctuations for TED-LIUM text automatically.
We test our methods on LibriSpeech test-clean set and TED-LIUM test set respectively. 

\subsubsection{E2E ASR model setting}
In our experiment, we use listen,attend and spell(LAS \cite{chan2015listen}) as ASR model.The encoder includes two convolutional layers and four bidirectional LSTM(BLSTM) layers. The first and second convolutional layers contain 64 and 128 filters with shape 3 x 3 and a 1x 1 stride,respectively.
The max-pooling following each convolutional layer downsample the input to half.
The decoder is a two unidirectional LSTM layers with 1024 units. 
We use location-based attention \cite{chorowski2015attention} and add SpecAugment \cite{48482}.
The optimization AdaDelta \cite{zeiler2012adadelta} is used to train ASR with gradient clipping is 5. 
When the accuracy of ASR on the validation set has not improved for five consecutive epochs, we stop training.
When decoding, we do not use any additional language models. The size of beam search is 20.

\subsubsection{Attention-based neural TTS model setting}
For our TTS model, we choose Tacotron2 \cite{shen2018natural,jia2018transfer}, which takes a text sequence as input, conditioned on speaker x-vector \footnote{x-vecoter is extracted from the speaker model shown in Figure \ref{fig1}, and its pre-trained model can be found in http://kaldi-asr.org/models/m8.} \cite{snyder2018x} to generate spectrogram for multi-speaker on-the-fly.
The input sequence embedding is encoded by three convolutional layers.
Every convolutional layer includes 512 filters with a shape 5x1.
A bidirectional long short-term memory (LSTM) layer following the convolutional layers has 256 units for each direction.
The optimization Adam \cite{kingma2014adam} is used to train TTS with gradient clipping is 1.
We use guided attention \cite{tachibana2018efficiently}  in baseline model training to get more stable alignment for long utterances.
\subsection{Results and Discussions}
\subsubsection{Domain adaptation for ASR}
In this experiment, we focus on mismatch of text in different domains and do adaptation using joint training of TTS and ASR as proposed in section 3. Table \ref{table:1} lists the results on different test sets.
For TED-LIUM test set, the WER of ASR baseline model is 25.9$\%$, which is much bigger than that on LibriSpeech test set. This is mainly due to the mismatch in content, speaker and environment. When we do domain adaptation via joint training with TTS, the WER is reduced to 23.3$\%$ on the TED-LIUM test set, about 10$\%$ relative reduction. While the performance is almost on par on the LibriSpeech test set after domain adaptation, 6.6$\%$ VS 6.5$\%$. It shows that our proposed method is effective at reducing mismatch in content of different domains. 
\begin{table}[th]
\centering
\caption{Performance of domain adaptation for LAS.}
\begin{tabular}{l||cc}
\hline
\multirow{2}{*}{Method} &  \multicolumn{2}{c}{WER(\%)}         \\ 
                        & LibriSpeech test set  &  TED test set  \\ \hline
Baseline                & 6.5                  & 25.9         \\ \hline 
\ \  +Domain adaptation      & 6.6                  & 23.3         \\ \hline 
\end{tabular}
\label{table:1}
\end{table}

We also compare different training strategies: 1) update both ASR and TTS during adaptation; 2) update only ASR and keep TTS fixed. The results are shown in Tabel \ref{table:2}. It can be seen that similar results can be obtained for both methods. This may be that updating TTS can't bring more acoustic or prosody diversity that is helpful for ASR on target domain, even though there is significant improvement in robustness, as we will shown below.

\begin{table}[h]
    \centering
    \caption{Comparison of different training strategies on TED-LIUM test set.}

    \begin{tabularx}{5cm}{llX}  
        \hline                      
        \multicolumn{1}{c}{Training Strategy}  & \multicolumn{1}{c}{WER(\%)} \\
        \hline
        \multicolumn{1}{c}{Update ASR} & \multicolumn{1}{c}{23.1}\\
        \multicolumn{1}{c}{Update ASR and TTS} & \multicolumn{1}{c}{23.3}\\
        \hline
    \end{tabularx}
    \label{table:2}
\end{table}

\subsubsection{Domain adaptation for TTS}
In our experiments, the TTS model is also updated during joint training for domain adaptation. We show the results of Tacotron2 in Table \ref{table:3}. Here, the baseline LAS model before adaptation is used to evaluate the synthesized speech by Tacotron2. It can be seen that significant improvements can be obtained in robustness of Tacotron2 for target domain (TED-LIUM test set), about 51.5$\%$ relative WER reduction in generated samples. Also, the stability of Tacotron2 in source domain is also improved, the WER of synthesized speech is reduced by about 16.9$\%$ relatively on LibriSpeech test set. It shows that joint training with ASR is very helpful in reducing attention errors for attention-based neural TTS.
\begin{table}[th]
\centering
\caption{Performance of domain adaptation for Tacotron2. The texts and utterances are randomly shuffled to generate (text, speaker reference) pairs for Tacotron2 input.}
\begin{tabular}{l||cc}
\hline
\multirow{2}{*}{Method} &  \multicolumn{2}{c}{WER(\%)}         \\ 
                        & LibriSpeech test set  &  TED test set  \\ \hline
Baseline                & 11.2                  & 23.1        \\ \hline 
\ \  +Domain adaptation      & 9.3                  & 11.2         \\ \hline 
\end{tabular}
\label{table:3}
\end{table}


We also show the improvement in robustness by plotting the alignment of input text and output mel-spectrogram generated by Tacotron2. From Figure \ref{fig2}, we can see that the alignment of all of three samples looks much better after joint training.

\subsubsection{Speaker adaptation for ASR}
We further apply few-shot speaker adaptation for LAS. In this experiment, we select a few utterances for each speaker in test set as speaker reference for Tacotron2 to generate speech. Here, the parameters of Tacotron2 are fixed during all training.

\begin{table}[th]
\caption{Performance of speaker adaptation for ASR on TED-LIUM.}
\centering
\begin{tabularx}{8cm}{llX}  
\hline                      
\multicolumn{1}{c}{Method}     & \multicolumn{1}{c}{\# of references}  & \multicolumn{1}{c}{WER(\%)} \\
\hline
\multicolumn{1}{c}{Domain adaptation}  & \multicolumn{1}{c}{0} & \multicolumn{1}{c}{23.1} \\
\multicolumn{1}{c}{\qquad +Speaker adaptation}  & \multicolumn{1}{c}{1} & \multicolumn{1}{c}{22.8} \\
\multicolumn{1}{c}{\qquad +Speaker adaptation}  & \multicolumn{1}{c}{5} & \multicolumn{1}{c}{22.5}\\
\hline
\end{tabularx}
\label{table:4}
\end{table}

The results for speaker adaptation are listed in Table \ref{table:4}. It can be seen that more gain can be obtained for LAS after speaker adaptation, and using more references is slightly better than using only one reference. This demonstrates that our proposed method is helpful in adapting to new speakers in a unsupervised way.

\section{Conclusions and Future work}
In this paper, we firstly investigate on domain adaptation using the TTS→ASR pipeline in machine speech chain. 
Our proposed method improves the performance of both ASR and TTS in the selected target domain, with nearly no degradation in the source domain.
We also compare different training strategies: one is joint training of TTS and ASR; the other is only updating ASR.
The latter result is slightly better in  SR evaluation, while the former can significantly improve the performance of TTS in the target domain.
Secondly, we apply the TTS→ASR pipeline on few-shot speaker adaptation for ASR.
Experiments show that speaker adaptation can further improve model performance.
In the future, we will combine our proposed method and ASR→TTS based on straight-through estimator to explore the unsupervised pre-training of ASR and TTS.

\label{sec:conclusion}


\bibliographystyle{IEEEtran}

\bibliography{mybib}


\end{document}

%% file: math_commands.tex
\usepackage{amsmath,amsfonts,bm}

\def\eqref#1{equation~\ref{#1}}

\def\1{\bm{1}}

\DeclareMathAlphabet{\mathsfit}{\encodingdefault}{\sfdefault}{m}{sl}
\SetMathAlphabet{\mathsfit}{bold}{\encodingdefault}{\sfdefault}{bx}{n}

